\documentclass[letterpaper]{article} 
\usepackage[submission]{aaai2026}  
\usepackage{times}  
\usepackage{helvet}  
\usepackage{courier}  
\usepackage[hyphens]{url}  
\usepackage{graphicx} 
\usepackage{natbib}  
\usepackage{caption} 
\usepackage{algorithm}
\usepackage{algorithmic}
\usepackage{booktabs}  
\usepackage{amsfonts}
\usepackage{amsmath}
\usepackage{amssymb}

\newcommand{\imageSpace}{\mathcal{I}}
\newcommand{\locationSpace}{\mathcal{L}}
\newcommand{\conceptSet}{\mathcal{C}}
\newcommand{\powerSet}[1]{\mathcal{P}(#1)}
\newcommand{\image}{I}
\newcommand{\location}{L}
\newcommand{\concept}[1]{c_{#1}}
\newcommand{\conceptSubset}{C_k}
\newcommand{\basisMatrix}{\mathbf{B}}
\newcommand{\textEmbedding}{\mathbf{E}_{\text{concept}}}
\newcommand{\offsetMatrix}{\Delta}
\newcommand{\imageFeatures}{\mathbf{x}_{\text{img}}}
\newcommand{\locationFeatures}{\mathbf{x}_{\text{loc}}}
\newcommand{\imageProjection}{\mathbf{z}_{\text{img}}}
\newcommand{\locationProjection}{\mathbf{z}_{\text{loc}}}
\newcommand{\imageEncoder}{E_{\image}}
\newcommand{\locationEncoder}{E_{\location}}

\newcommand{\mlp}{f_{\text{img}}}

\usepackage{newfloat}
\usepackage{listings}
\DeclareCaptionStyle{ruled}{labelfont=normalfont,labelsep=colon,strut=off} 
\floatstyle{ruled}
\newfloat{listing}{tb}{lst}{}
\floatname{listing}{Listing}
\title{Towards Interpretable Geo-localization: a Concept-Aware Global Image-GPS Alignment Framework}

\author{
  Furong Jia\equalcontrib \textsuperscript{\rm 1},
  Lanxin Liu\equalcontrib \textsuperscript{\rm 2},
  Ce Hou \textsuperscript{\rm 3},
  Fan Zhang\textsuperscript{\rm 1}\thanks{Corresponding author: \texttt{fanzhanggis@pku.edu.cn}},
  Xinyan Liu \textsuperscript{\rm 4},
  Yu Liu \textsuperscript{\rm 1} \\
  \normalfont
  \textsuperscript{\rm 1}Peking University \\
  \textsuperscript{\rm 2}Harbin Institute of Technology \\
  \textsuperscript{\rm 3}The Hong Kong University of Science and Technology \\
  \textsuperscript{\rm 4}Harbin Institute of Technology (Weihai) \\
}

\usepackage{bibentry}
\def\showauthors@on{T}

\begin{document}

\maketitle

\begin{abstract}
Worldwide geo-localization involves determining the exact geographic location of images captured globally, typically guided by geographic cues such as climate, landmarks, and architectural styles. Despite advancements in geo-localization models like GeoCLIP, which leverages images and location alignment via contrastive learning for accurate predictions, the interpretability of these models remains insufficiently explored.
Current concept-based interpretability methods fail to align effectively with Geo-alignment image-location embedding objectives, resulting in suboptimal interpretability and performance.
To address this gap, we propose a novel framework integrating global geo-localization with concept bottlenecks. 
Our method inserts a Concept-Aware Alignment Module that jointly projects image and location embeddings onto a shared bank of geographic concepts (e.g., tropical climate, mountain, cathedral) and minimizes a concept-level loss, enhancing alignment in a concept-specific subspace and enabling robust interpretability.
To our knowledge, this is the first work to introduce interpretability into geo-localization.
Extensive experiments demonstrate that our approach surpasses GeoCLIP in geo-localization accuracy and boosts performance across diverse geospatial prediction tasks, revealing richer semantic insights into geographic decision-making processes.
\end{abstract}



\section{Introduction}
Image geo-localization refers to the task that determine
geographic coordinates from visual content alone.
With applications ranging from 
ecological monitoring~\cite{beery2022auto,russwurm2020meta} to disaster response~\cite{sathianarayanan2024extracting}, it has attracted increasing attention.
However, achieving accurate worldwide image geo-localization remains technically challenging since the enormous variability in Earth's geographical landscapes.

\begin{figure}[t]
\centering
\includegraphics[width=0.95\columnwidth]{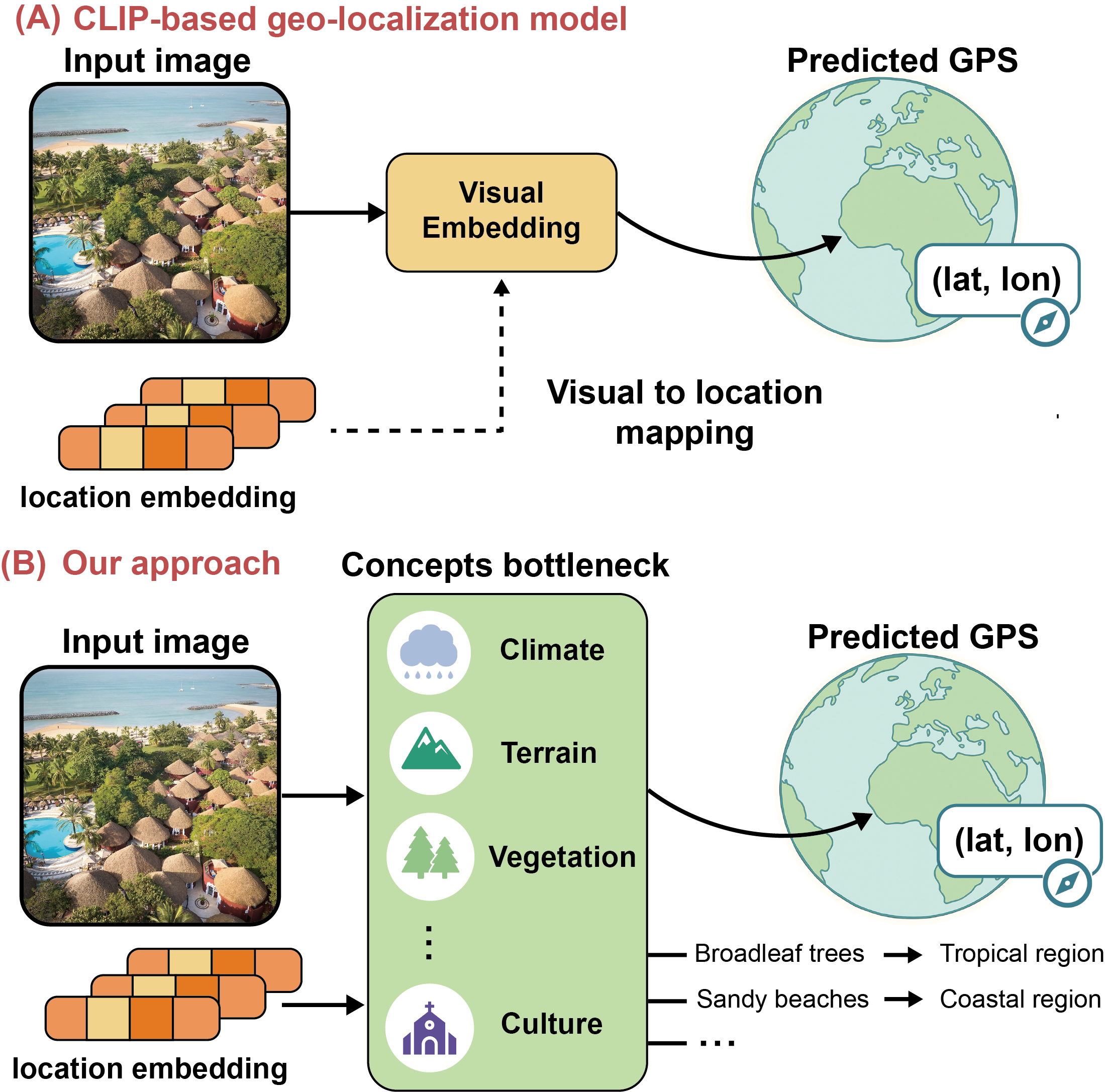} 
\caption{
\textbf{Concept-aware geo-localization overview.}
(A) CLIP-based geo-localization model predicts GPS coordinates by directly mapping visual embeddings to a location embedding gallery, following the image-location alignment.
(B) Our approach inserts a concept bottleneck between image and GPS, grounding GPS prediction in semantically meaningful cues.
}
\label{title-image}
\end{figure}

Recent CLIP-based geo-localization models (Fig.~\ref{title-image}A) ~\cite{vivanco2023geoclip, klemmer2025satclip, zhong2022regionclip} have shown promising performance in zero-shot geo-localization by aligning image and geographical location (latitude, longitude) embeddings in a shared space. 
While this approach enables global location retrieval, it still faces the challenge of distinguishing visually similar scenes from geographically distinct areas. This is because successful geo-localization requires more than just learning spatial distributions of visual landscapes, it demands an understanding of broader geographic knowledge and fine-grained concepts (Fig.~\ref{title-image}B) to distinguish between locations across the globe. Relying solely on raw geographical coordinates to learn location embeddings is insufficient for encoding such world knowledge.

Beyond performance, interpretability of image geo-localization is also important not only for reliable geographic decision-making but also for understanding the spatial cognitive abilities of AI~\cite{roberts2023gpt4geo}.
The reasoning behind geo-localization is inherently complex, involving diverse cues like landmarks, architectural style, and environmental features. 
While prior work either overlooks interpretability or focuses solely on pixel-level attribution methods~\cite{theiner2022interpretable} that quantify region importance through low-level visual features, it lacks the conceptual understanding required for high-level tasks.
This underscores the need for concept-based interpretability that grounds predictions in human-understandable semantic concepts.

To achieve more precise and interpretable geo-localization, we introduce a framework that integrates precise location recognition with concept-driven alignment.
Specifically, we introduce a Geography-Driven Concept Set, comprising regionally significant attributes carefully selected for their clear geospatial meaning, allowing the model to ground its predictions in recognizable geographic features. This enhances the model's interpretability by linking its outputs to well-defined concepts that are easily understood.
Additionally, we propose a Concept-Aware Alignment Module that aligns image and location embeddings within a geography-driven concept subspace defined by CLIP-based embeddings. 
The concept-aware alignment encourages image and location embeddings to converge toward similar distributions within an interpretable semantic space, enhancing interpretability, while preserving accuracy.

Our contributions can be summarized as follows:
\begin{itemize}
    \item We develop a systematic approach to construct Geography-Driven Concept Sets from regionally salient attributes for interpretable geographic grounding.
    \item We propose a Concept-Aware Alignment Module that aligns image and location embeddings within an interpretable, geography-driven concept subspace.
    \item Extensive experiments demonstrate that our framework not only provides intuitive explanations but also improves GeoCLIP's performance on geo-localization tasks as well as downstream tasks.
\end{itemize}


\section{Related Work}
\subsection{Image Geo-localization}
While recent studies have explored the potential of large language models (LLMs) for geo-localization via natural-language reasoning and direct coordinate prediction~\cite{xu2024addressclip,jia2024g3,zhou2024img2loc,jia2025georanker,ye2024cross,wang2025gre}, the majority of existing methods still fall into two primary paradigms: classification-based and retrieval-based approaches~\cite{vivanco2023geoclip}.

Classification-based approaches discretize the earth's surface into predefined cells and train models to assign each image to one of these regions~\cite{weyand2016planet, seo2018cplanet,vo2017revisiting,pramanick2022world,clark2023we}.
While conceptually straightforward, this formulation enforces a fixed spatial granularity that often causes large localization errors near cell boundaries or in underrepresented areas.
To improve interpretability under this paradigm, Theiner et al.~\cite{theiner2022interpretable} introduce a semantic partitioning strategy that replaces fixed or arbitrarily defined spatial partitions with more coherent region definitions.

Another widely adopted paradigm is retrieval-based geo-localization, where the goal is to match a query image against a gallery—either of reference images or location embeddings—based on similarity~\cite{workman2015wide,tian2017cross,liu2019lending,shi2020looking,yang2021cross,zhu2023r2former}.
Among retrieval-based methods, GeoCLIP was the first work employing geographical location encoding and retrieve location through geo-alignment~\cite{vivanco2023geoclip}, leveraging contrastive paradigm and CLIP’s pretrained image encoder~\cite{radford2021learning}. 
Building on this, we propose enhancing image–location retrieval with interpretable geographic concepts as intermediate anchors for alignment.

\subsection{Concept-based Model Interpretation }

The concept-based interpretation has received considerable attention for its ability to ground model decisions in high-level, human-interpretable concepts~\cite{yeh2020completeness}.
This approach is particularly important in image classification~\cite{ghorbani2019towards} and multimodal learning~\cite{parekh2024concept}, 
where providing transparent rationales for predictions is essential for building user trust.

The Concept Bottleneck Model (CBM)~\cite{koh2020concept} exemplifies this paradigm. CBMs decompose prediction into two sequential stages: first, inferring intermediate concepts from inputs, and then using these inferred concepts to produce the final classification. 
By employing unsupervised techniques to discover and utilize latent concepts, label-free CBMs~\cite{oikarinenlabel} expand the concept-based paradigm to settings where labeled concept annotations are unavailable.
Post-hoc CBMs~\cite{yuksekgonulpost} enable interpretability in pre-trained models without changing their original parameters.
Recently, language models have demonstrated the ability to define and organize concepts in a way that aligns with human cognition, enhancing the usability of concept-based interpretability frameworks~\cite{yang2023language}.
This line of work is further advanced by models that eliminate the need for predefined concept sets, instead generating fluent and accurate natural language explanations directly from inputs using pretrained vision–language models~\cite{yamaguchi2025explanation}.

\begin{figure*}[t]
\centering
\includegraphics[width=0.99\textwidth]{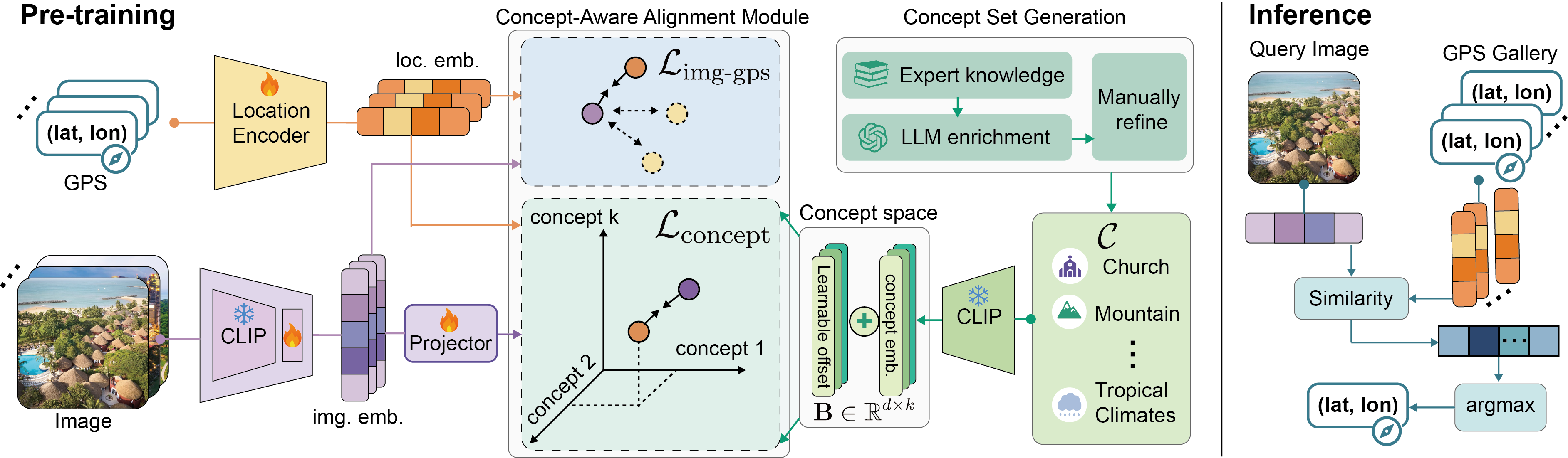} 
\caption{
\textbf{The pretraining and inference pipeline.}
\textit{Left:} The pretraining framework with a Concept-Aware Alignment Module integrated into the image–GPS contrastive learning process.
\textit{Right:} The inference process of image geo-localization.
}
\label{backbone}
\end{figure*}

\section{Method}

\subsection{Definition of Interpretability for Geo-localization}
We regard image-based geo-localization as a concept-guided retrieval problem, where the predicted location corresponds to the geographical coordinate whose embedding is closest to that of the input image.
Given an image \(\image \in \imageSpace\), our aim is to learn a mapping
$
F: \imageSpace \longrightarrow \locationSpace \times \powerSet{\conceptSet},
$
where \(\locationSpace\) denotes the continuous space of geographical coordinates (latitude--longitude pairs) and \(\powerSet{\conceptSet}\) the set of a predefined concept library \(\conceptSet = \{\concept{1}, \ldots, \concept{n}\}\). For each input, \(F\) produces a predicted location \(\location \in \locationSpace\).

The overall architecture is illustrated in Figure~\ref{backbone}. 
To enable concept-guided geolocation, our framework extracts and aligns features from images, GPS locations, and geographic concepts. 
We integrates a \textbf{Geography-Driven Concept Set} \(\mathcal{C} = \{c_1, \dots, c_n\}\), where each \(c_i\) represents a regionally salient attribute (e.g., landform, architectural style) with explicit geospatial semantics, alongside a \textbf{Concept-Aware Alignment Module} \(M(\cdot)\). We employ CLIP-based encoders for all three modalities: an image encoder \(E_{\mathrm{img}}\) that maps an image \(I\) to features \(x_{\mathrm{img}} = E_{\mathrm{img}}(I) \in \mathbb{R}^d\), a location encoder \(E_{\mathrm{loc}}\) that transforms GPS coordinates \(L \in \mathcal{L}\) into \(x_{\mathrm{loc}} = E_{\mathrm{loc}}(L) \in \mathbb{R}^d\) (cf. GeoCLIP~\cite{vivanco2023geoclip}) and a text encoder \(E_{\mathrm{text}}\) that maps concept text into concept text embeddings. 
Within \(M(\cdot)\), we align \(x_{\mathrm{img}}\) and \(x_{\mathrm{loc}}\) using contrastive learning in the original feature space, then project both into a concept subspace via a learnable basis matrix \(B \in \mathbb{R}^{d \times k}\). This dual-alignment approach enhances localization accuracy while grounding predictions in interpretable geographic concepts.

\subsection{Construction of Geography-Driven Concept Set}
We construct a concept set \(\conceptSet\) that captures geographically meaningful semantics,
following a three-step 
procedure:

(1) Grounding in geographic knowledge: We first construct a comprehensive pool of geographically relevant concepts by extracting domain-specific expert knowledge from extensive textual resources, such as Wikipedia and domain-specific knowledge graphs (e.g., WorldKG~\cite{dsouza2021worldkg}). These sources encompass both natural geographic knowledge (e.g., climate types) and human geographic knowledge (e.g., transportation), ensuring that our initial concept set is extensive and geographically informed.

(2) LLM-based enrichment: Recent research has validated the effectiveness of LLMs in generating robust and contextually nuanced visual concepts~\cite{ruiz2024theoretical}.
To enrich the initial concept set and capture finer-grained and culturally distinctive elements, we further utilize LLMs such as GPT4~\cite{achiam2023gpt}.
The prompt we used is included in the supplementary material.


(3) Manual refinement: Finally, we manually review and filter the generated concepts to achieve conciseness and effectiveness. 



\subsection{Concept-Aware Alignment Module}
We first construct a concept basis from the concept set \(\conceptSet\), where concepts are encoded using a frozen CLIP text encoder to obtain a matrix of semantic directions. A learnable offset is added to tailor the concept basis to the image geolocalization task, yielding a tunable set of concept vectors \(\basisMatrix \in \mathbb{R}^{d \times k}\), where \(k\) is the number of selected concepts in the subset \(\conceptSubset \subseteq \conceptSet\). Formally, the concept basis is constructed as:
\begin{equation}
\basisMatrix = \textEmbedding + \offsetMatrix,
\end{equation}
where \(\textEmbedding \in \mathbb{R}^{d \times k}\) denotes the fixed textual embeddings of the \(k\) concepts in \(\conceptSubset\) obtained from a frozen CLIP text encoder \(E_{\mathrm{text}}\), and \(\offsetMatrix \in \mathbb{R}^{d \times k}\) is a learnable offset matrix optimized during training to adapt the concept directions to the downstream geolocalization task.

During training, the image features \(\imageFeatures = \imageEncoder(\image) \in \mathbb{R}^{d}\) are projected into the concept subspace via a lightweight MLP:
\begin{equation}
\imageProjection = \mlp(\imageFeatures) \in \mathbb{R}^{k},
\end{equation}
where \(\mlp(\cdot)\) denotes the MLP, and \(k\) is the number of selected concepts in \(\conceptSubset\).

In parallel, the location features \(\locationFeatures = \locationEncoder(\location) \in \mathbb{R}^{d}\) are directly projected using the concept basis matrix \(\basisMatrix \in \mathbb{R}^{d \times k}\):
\begin{equation}
\locationProjection = \locationFeatures^{\top} \basisMatrix \in \mathbb{R}^{k}.
\end{equation}

Both \(\imageProjection\) and \(\locationProjection\) lie in the interpretable concept subspace and serve as semantic representations aligned across modalities.

\subsubsection*{Loss Function}

Our objective combines a cross-modal alignment loss and a distribution-level consistency loss in the concept space.

First, we define the image-to-location contrastive loss as a standard cross entropy loss over the similarity logits. Given a batch of $N$ image-location pairs $\{(\mathbf{z}^\text{img}_i, \mathbf{z}^\text{loc}_i)\}_{i=1}^{N}$, the loss is computed as:

\begin{equation}
\mathcal{L}_{\text{img-gps}} = - \frac{1}{N} \sum_{i=1}^{N} \log \frac{
\exp\left( {\mathbf{z}^\text{img}_i \cdot \mathbf{z}^\text{loc}_i}/{\tau} \right)
}{
\sum_{j=1}^{N} \exp\left( {\mathbf{z}^\text{img}_i \cdot \mathbf{z}^\text{loc}_j}/{\tau} \right)
},
\end{equation}
where $\tau$ is a learnable temperature parameter that controls the scale of the similarity scores. This objective encourages each image representation to be most similar to its corresponding GPS embedding while pushing apart unrelated pairs.

To encourage modality-invariant concept representations, we further introduce a Concept Space Divergence Loss, inspired by multi-modal distributional alignment techniques~\cite{yin2025distributionalvisionlanguagealignmentcauchyschwarz}. Specifically, we use a Gaussian kernel function:
\begin{equation}
K(\mathbf{x}, \mathbf{y}) = \exp\left(-\frac{\|\mathbf{x} - \mathbf{y}\|^2}{2\sigma^2}\right),
\end{equation}
and compute the divergence between the projected image features $\mathbf{z}_\text{img}$ and projected location features $\mathbf{z}_\text{loc}$ as:
\begin{equation}
\label{eq}
\begin{aligned}
    \mathcal{L}_{\mathrm{concept}}
    = \frac{1}{N^2}
    \sum_{i,j=1}^{N}
    \Bigl[
        \log K\bigl(\mathbf{z}^{\mathrm{img}}_{i}, \mathbf{z}^{\mathrm{img}}_{j}\bigr) 
        &+ \log K\bigl(\mathbf{z}^{\mathrm{loc}}_{i}, \mathbf{z}^{\mathrm{loc}}_{j}\bigr) \\
        - 2\,\log K\bigl(\mathbf{z}^{\mathrm{img}}_{i}, \mathbf{z}^{\mathrm{loc}}_{j}\bigr)
    \Bigr].
\end{aligned}
\end{equation}

The total loss is given by:
\begin{equation}
\mathcal{L} = \mathcal{L}_{\text{img-gps}} + \lambda \cdot \mathcal{L}_{\text{concept}},
\end{equation}
where $\lambda$ is a weighting coefficient.


\section{Experiments}

We structure our experiments around four research questions that investigate how to enhance the interpretability and performance of multimodal contrastive learning approaches for image geo-localization.
First, we explore whether the introduction of Concept-Aware Alignment Module
contributes to overall performance, both in terms of \textit{image geo-localization task (RQ 1)} and other \textit{downstream geospatial tasks (RQ 2)}. 
To this end, we evaluate the pretrained concept-aware embeddings on a set of downstream geospatial tasks, comparing their effectiveness with baseline contrastive learning-based models that do not leverage concept bottlenecks.

To explore the interpretability of image geo-localization,
we further ask: to what extent can \textit{the interpretability results reveal geographically meaningful patterns that align with human understanding (RQ 3)}.
We examine whether the explanations derived from our model, at both the individual and global levels, reveal established geographic knowledge, including both natural and cultural features.
Finally, we explore \textit{whether concept-aware contrastive pretraining facilitates the emergence of more geographically structured embeddings for both images and locations (RQ 4)}.

In the remainder of this section, we describe our experimental setup in detail, including datasets, downstream tasks. 

\subsection{Datasets and Implementation Details}
We train our model using the MediaEval Placing Task 2016 (MP-16)~\cite{larson2017benchmarking} dataset, a curated subset of the Yahoo Flickr Creative Commons 100 Million (YFCC100M)~\cite{thomee2016yfcc100m}. For evaluation, we test the trained model on the Im2GPS3k~\cite{hays2008im2gps} benchmark, following GeoCLIP~\cite{vivanco2023geoclip}. Additional datasets used for downstream tasks, along with task definitions, are detailed in the following section. 

To ensure training efficiency, we pretrain our model on a 5\% randomly sampled subset of the MP-16 dataset. The model is trained using the Adam optimizer with learning rates of $3 \times 10^{-5}$ for the location encoder and $3 \times 10^{-4}$ for other components. The hyperparameter $\lambda$ in Eq.~\ref{eq} is set to 10, and the batch size is 128. Training is conducted on an NVIDIA A6000 GPU.

In evaluation, similar to GeoCLIP~\cite{vivanco2023geoclip}, we also uses a Ten Crop strategy, where predictions from five distinct image crops and their horizontal flips are aggregated through averaging to produce the final prediction.
Performance is reported using a threshold-based metric, where we compute the geodesic distance between predicted and ground-truth locations and calculate the percentage of predictions within predefined distance thresholds (1 km, 25 km, 200 km, 750 km, and 2500 km).

\subsection{Downstream Tasks}
To investigate whether pretraining with a concept bottleneck enhances the expressiveness of location embeddings, we evaluate their performance on geospatial downstream tasks beyond image geo-localization.
We use the pretrained location embeddings to predict geographic attributes, including socioeconomic characteristics and environmental factors, based on input geographical coordinates, using multi-layer perceptron (MLP) models.
For socioeconomic attributes, 
we construct a nationwide dataset of \textbf{median income} and \textbf{educational attainment} at the census tract level using data from U.S. Census Bureau data. 
The educational attainment task involves predicting the proportion of the population with a \textbf{Bachelor's degree} and a \textbf{Graduate degree}.
We additionally include a \textbf{country} classification task~\cite{klemmer2025satclip} as part of the evaluation.
For environmental factors, we perform an \textbf{air temperature} prediction task~\cite{hooker2018global} and a \textbf{species} classification task~\cite{van2018inaturalist}.
In the species classification task, we extract image embeddings from our visual encoder and concatenate them with the location embeddings.

All downstream tasks use MSE loss for regression and cross-entropy for classification. Hyperparameters such as learning rate, number of layers, and hidden dimensions are tuned via random search on a validation set, and results are reported on a held-out test set.


\subsection*{\hspace{0pt}\makebox[\linewidth][l]{\textbf{Performance on Image Geo-localization} \hfill \textit{(RQ 1)}}}

\begin{table}[ht]
\centering
\caption{
Our method outperforms GeoCLIP and other geo-localization methods on Im2GPS3k, achieving consistent improvements across different distance thresholds.
}
\label{tab:geoloc_accuracy}
\small
\setlength{\tabcolsep}{3.5pt}   
\begin{tabular}{@{}l|c|c|c|c|c@{}}
\toprule
\textbf{Method} & \textbf{Street} & \textbf{City} & \textbf{Region} & \textbf{Country}  & \textbf{Continent} \\
 & \textit{(1 km)} & \textit{(25 km)} & \textit{(200 km)} & \textit{(750 km)} & \textit{(2500 km)} \\
\midrule
{[L]}kNN, $\sigma =4$ & 7.2 & 19.4 & 26.9 & 38.9 & 55.9 \\
PlaNet & 8.5 & 24.8 & 34.3 & 48.4 & 64.6 \\
CPlaNet & 10.2 & 26.5 & 34.6 & 48.6 & 64.6 \\
ISNs & 3.2 & 9.6 & 14.3 & 25.1 & 43.9 \\
Translocator & 7.6 & 20.3 & 27.1 & 40.7 & 63.3 \\
GeoDecoder & 5.7 & 10.3 & 21.4 & 28.9 & 38.6 \\
GeoCLIP & 10.8 & 31.1 & 48.7 & 67.6 & 83.2 \\
\textbf{Ours} & \textbf{13.2} & \textbf{34.0} & \textbf{49.8} & \textbf{68.2} & \textbf{83.5} \\
\bottomrule
\end{tabular}
\end{table}


We conduct a comparative evaluation of our model on global image geo-localization benchmarks. Specifically, we compare against the following methods:
kNN\cite{vo2017revisiting},
PlaNet\cite{weyand2016planet},
CPlaNet\cite{seo2018cplanet},
ISNs\cite{muller2018geolocation},
Translocator\cite{pramanick2022world},
GeoDecoder\cite{clark2023we}, and
GeoCLIP~\cite{vivanco2023geoclip}.

Table~\ref{tab:geoloc_accuracy} presents the results on the Im2GPS3k dataset. Across all distance thresholds, our model consistently outperforms GeoCLIP and other baselines, with improvements of +2.4\%, +2.9\%, +1.1\%, +0.6\%, and +0.3\% at the 1km, 25km, 200km, 750km, and 2500km levels, respectively. 
The comparison is conducted using 5\% of the full MP-16 dataset~\cite{wang2025gre}.

\subsection*{\hspace{0pt}\makebox[\linewidth][l]{\textbf{Downstream Task Performance} \hfill \textit{(RQ 2)}}}
As shown in Table~\ref{tab:downstream}, our model consistently outperforms GeoCLIP on both the socioeconomic attribute prediction and environmental factor prediction tasks. These tasks span different spatial scales, ranging from nationwide to global coverage. The results demonstrate that concept-aware pretraining enhances the expressiveness and utility of location embeddings across diverse geographic prediction tasks.

\begin{table}[t]
\small
\centering
\setlength{\tabcolsep}{2pt}      
\begin{tabular}{lccc}
\toprule
\textbf{Task $\downarrow$ Data $\rightarrow$} & 
\textbf{Spatial coverage} &
\textbf{Ours} &
\textbf{GeoCLIP} \\
\midrule
\multicolumn{3}{l}{\textbf{Regression} } & $R^{2}\! \uparrow$ \\
Air temperature & Global & \textbf{0.7538} & 0.7257 \\
Median income  & USA & \textbf{0.5468} & 0.4983 \\
Bachelor's degree  & USA & \textbf{0.5386} & 0.4742 \\
Graduate degree  & USA & \textbf{0.5524} & 0.5123 \\
\midrule
\multicolumn{3}{l}{\textbf{Classification}} & Accuracy $\uparrow$ \\
Countries  & Global & \textbf{91.12} & 90.72 \\
iNaturalist & Global & \textbf{65.94} & 62.01  \\
\bottomrule
\end{tabular}
\caption{Performance comparison between our model and GeoCLIP on downstream tasks.}
\label{tab:downstream}
\small
\end{table}

\subsection*{\hspace{0pt}\makebox[\linewidth][l]{\textbf{Interpretable Geo-localization} \hfill \textit{(RQ 3)}}}

\begin{figure}[t]
\centering
\includegraphics[width=0.95\columnwidth]{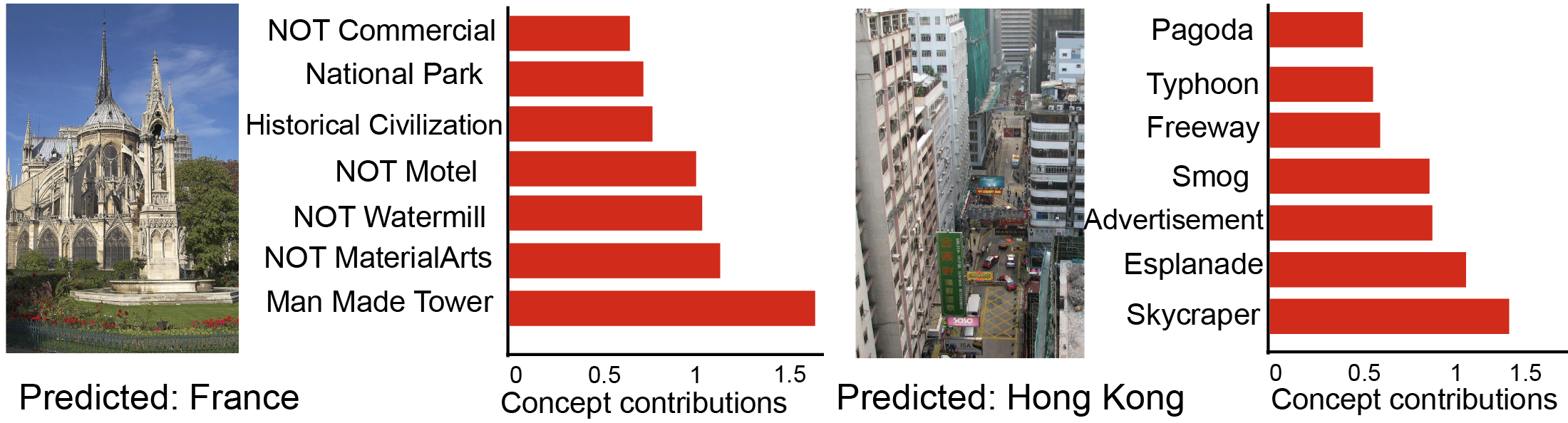} 
\caption{\textbf{Visualization of concept contributions} in country-level geo-localization task with Label-free CBM.}
\label{fig-cbm}
\end{figure}

\subsubsection{Individual 
explainable cases}
To probe the interpretability of the representations learned with our Concept-Aware Image–GPS Contrastive Module, we cast a country-level classification task that takes the image features produced by our pretrained image encoder as input.
A Label-Free CBM ~\cite{oikarinenlabel} is trained on these features; its final normalized linear weights serve as concept-contribution scores. 
Figure ~\ref{fig-cbm} illustrates two correctly classified examples (more visualizations are available in the supplementary material).
The model assigns the highest positive weights to the concepts ``skyscraper" and ``esplanade," both salient visual cues of Hong Kong. It also highlights ``advertisement" and the climate-related concept ``typhoon," consistent with the city’s dense commercial signage and subtropical weather.
For the French scene, the concept ``historical civilization", triggered by the prominent heritage architecture, dominates the decision, aligning well with human intuition.


\begin{figure}[t]
\centering
\includegraphics[width=0.95\columnwidth]{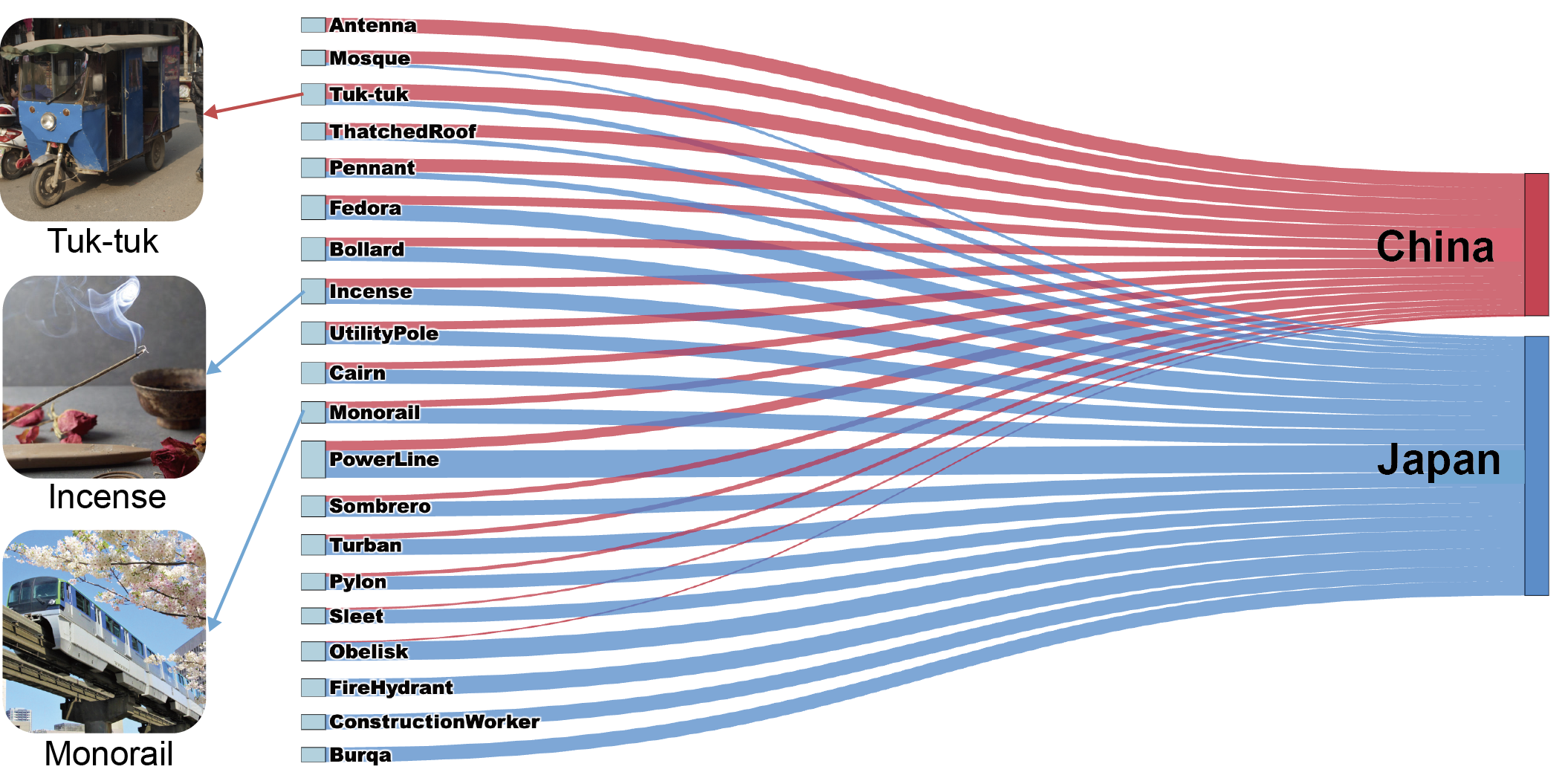} 
\caption{
\textbf{Global decision rules in image geo-localization.}
Diagram showing prominent visual concepts differentiating image geo-localization predictions between China and Japan. Examples highlight region-specific concepts illustrating how the model's learned concepts reflect meaningful geographic and cultural patterns. Zoom in for better view.}
\label{sankey-image}
\end{figure}

\subsubsection{Global decision rules}
We further examine whether our concept-aware model captures global decision rules that are interpretable at an aggregated level. 
In Figure~\ref{sankey-image}, 
we show identified prominent concepts that distinguish China and Japan by measuring per-concept activation levels in the concept subspace of our Concept-Aware Image–GPS Contrastive Module.
The Sankey diagram clearly reveals concepts that align well with intuitive geographic and cultural understanding. For example, ``Tuk-tuk" is distinctly prevalent in China, reflecting its popularity due to versatility and widespread usage. ``Incense," notably connects with the traditional Japanese tea ceremony (Chado), emphasizing its cultural resonance across regions. Similarly, ``Monorail," while existing in both countries, is strongly associated with Japan due to landmarks like the historically significant Shonan Monorail. 
The results suggest that the learned concepts effectively capture geographically distinctive features,
shedding light on the global interpretability of geo-localization model.

\subsubsection{Concept-level activation}

\begin{table*}[ht]
\setlength{\tabcolsep}{3pt}
\centering
\caption{
Top-8 and lowest-8 concepts by median score (ms) across distance error intervals.
}
\label{tab:concept-influence}
\small
\begin{tabular}{@{}c c | c c | c c || c c | c c | c c@{}}
\toprule
\multicolumn{6}{c||}{Top-8} & \multicolumn{6}{c}{Lowest-8} \\
\cmidrule{1-12}
\multicolumn{2}{c|}{[0--25)} & \multicolumn{2}{c|}{[25--200)} & \multicolumn{2}{c||}{[200--750)} & \multicolumn{2}{c|}{[0--25)} & \multicolumn{2}{c|}{[25--200)} & \multicolumn{2}{c}{[200--750)} \\
\cmidrule{1-12}
concept & ms & concept & ms & concept & ms & concept & ms & concept & ms & concept & ms \\
\midrule
Windmill & 0.1117 & Eucalyptus & 0.1139 & Eucalyptus & 0.1188 & Island & 0.0561 & Motorcycle & 0.0685 & Cobblestone & 0.0673 \\
Theater & 0.1085 & Cairn & 0.1073 & PowerLine & 0.1108 & Uniform & 0.0547 & Fortress & 0.0678 & Souk & 0.0662 \\
Factory & 0.1067 & PrayerFlags & 0.1067 & Crater & 0.1084 & Tea & 0.0533 & Hedge & 0.0660 & Woodland & 0.0636  \\
Scaffolding & 0.1054 & Grapevine & 0.1060 & Favela & 0.1063 & Brickwork & 0.0527 & Townhouse & 0.0640 & Oak Tree & 0.0630 \\
Promenade & 0.1052 & Favela & 0.1060 & Vending & 0.1024 & Reflection & 0.0489 & Hairstyle & 0.0627  & Shanty town& 0.0602 \\
Drizzle & 0.1041 & Temperate & 0.1059 & Chimney & 0.1016 & Gate & 0.0454 & Succulent & 0.0627  & Summit& 0.0592 \\
Mangrove & 0.1041 & Sari & 0.1052 & Dune & 0.1011 & Lorry & 0.0449 & SandDune & 0.0623  & Pennant& 0.0519 \\
Sombrero & 0.1037 & Underpass & 0.1045 & Gondola & 0.1009 & Bus & 0.0448 & Plantation & 0.0621 & Demonstration & 0.0472 \\
\bottomrule
\end{tabular}
\end{table*}

To assess the relative activation levels of individual concepts in our model's representations and predictions, we analyze concept-wise scores on the Im2GPS3k dataset. Each concept score is derived from the image's projection in the concept subspace, where the projected value represents the activation strength of that concept. We enforce sparsity by retaining only the top-20 scoring concepts per image. Then, we compute the median score for each concept across all images and rank them within three geolocation error intervals. As shown in Table~\ref{tab:concept-influence}, we report the top-8 and bottom-8 concepts with the highest and lowest median scores (ms), respectively, within each spatial bin.
Although these activation values do not directly indicate causal influence, higher concept activation may suggest a stronger presence or reliance of that concept in the prediction process. 

\subsection*{\hspace{0pt}\makebox[\linewidth][l]{\textbf{Concept-Aware Embedding Analysis} \hfill \textit{(RQ 4)}}}
\begin{figure}[t]
\centering
\includegraphics[width=0.9\columnwidth]{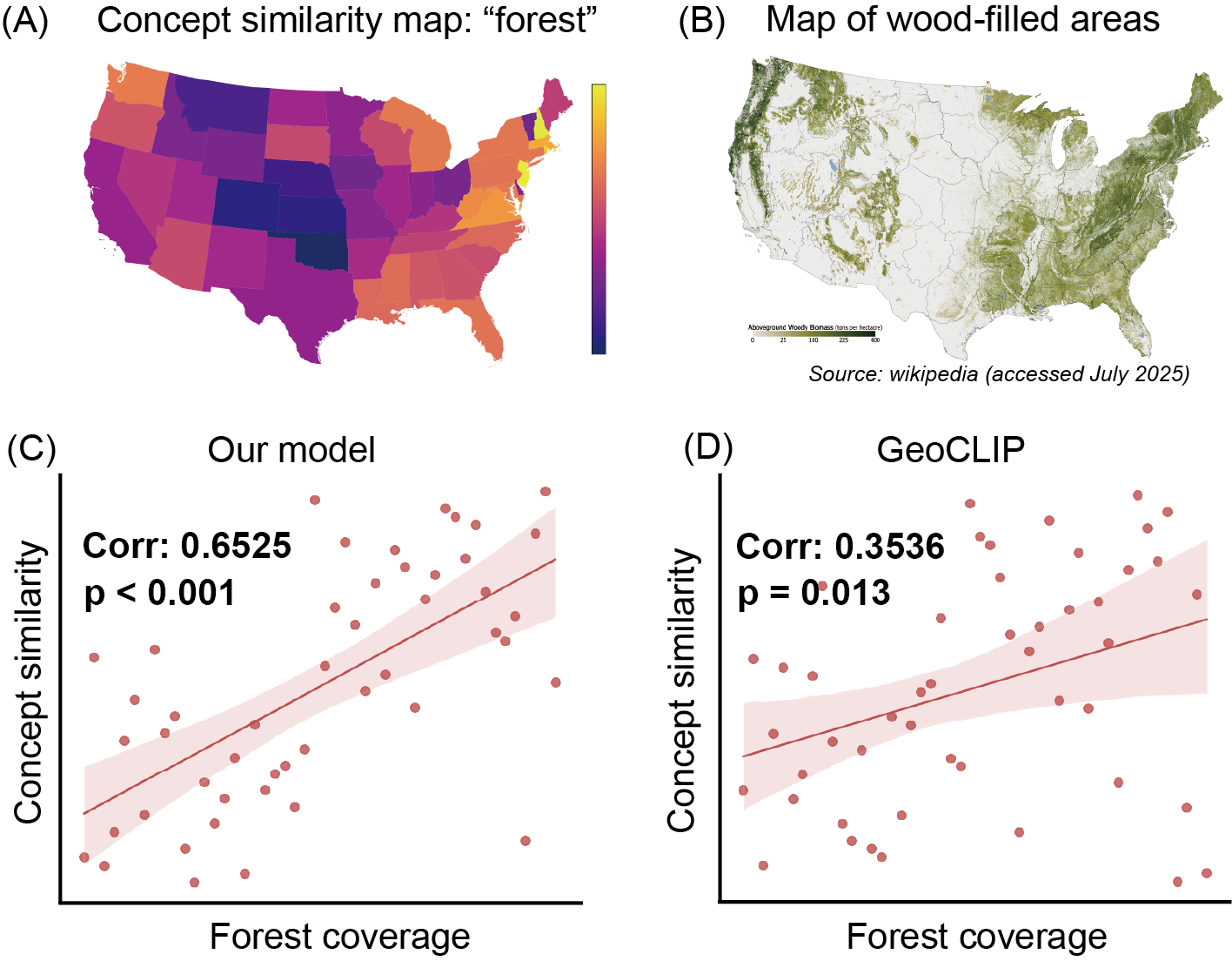} 
\caption{
\textbf{Visualization of how location embeddings capture interpretable geographic concepts aligned with real-world distributions.}
(A) State-level similarity map for the concept ``forest” derived from our model’s location embeddings. (B) Reference map of forests across the U.S. (C) Correlation between actual forest coverage and concept similarity from our model. (D) Correlation using concept similarity from GeoCLIP.}
\label{fig-concept-forest}
\end{figure}

We demonstrate the concept bottleneck enhances geographic semantics in location embeddings.
Since the location encoder has been trained to align with the pre-trained CLIP text embedding space, we can create concept similarity maps by measuring the similarity between each location embedding and a given concept's text embedding~\cite{vivanco2023geoclip}.
For example, by querying the concept ``forest”, we generate a state-level similarity map across the U.S. (Figure~\ref{fig-concept-forest}A), where the value for each state is obtained by averaging the similarity across sample points within the state. As shown in Figure~\ref{fig-concept-forest}B, which presents a reference map of forests in the U.S., the highlighted regions in the similarity map corresponds to known forest-dense regions.

Quantitatively, we assess the alignment between the concept similarity and the ground truth geospatial distribution by computing the Pearson correlation coefficient ($\rho$) at the U.S. state level. We use 2016 forest coverage data obtained from Wikipedia  
as the reference.
As shown in Figure~\ref{fig-concept-forest}C and Figure~\ref{fig-concept-forest}D, our model's similarity map achieves a higher correlation ($\rho = 0.6525$, $p < 0.001$) compared to GeoCLIP ($\rho = 0.3536$, $p = 0.013$), indicating enhanced representation of geographic concepts in our location embeddings.

\begin{figure}[t]
\centering
\includegraphics[width=0.85\columnwidth]{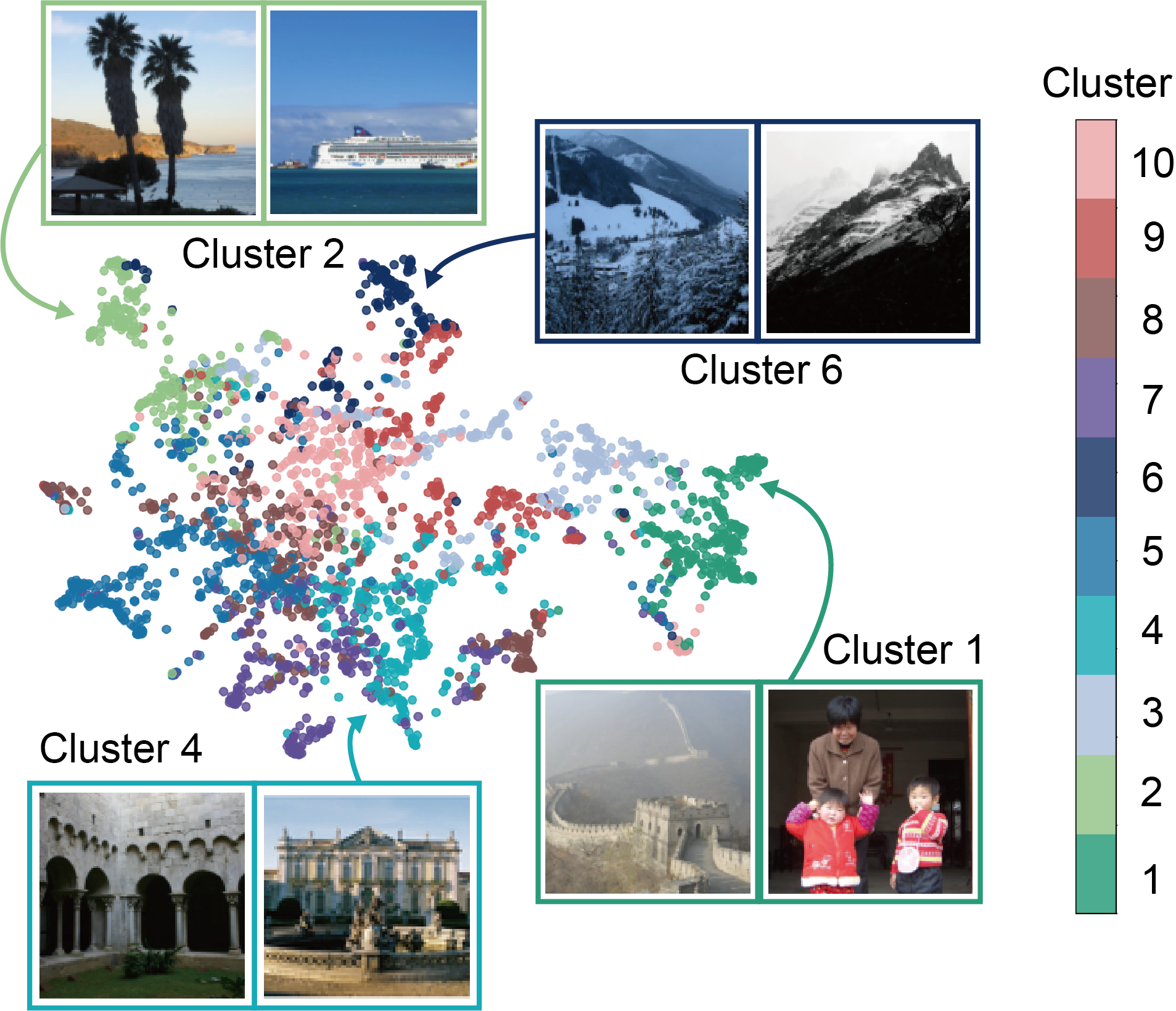} 
\caption{
\textbf{UMAP visualization of image embeddings} from the Im2GPS3k dataset learned by the Concept-Aware Alignment Module. Each point represents an image colored by k-means cluster assignment. More visualizations are provided in the supplementary materials.}
\label{fig-image-emb}
\end{figure}

To further answer \textit{RQ4} and explore whether concept-aware pretraining alters how image representations encode geographic information,
we apply dimensionality reduction (UMAP) to visualize image embeddings from Im2GPS3k.
We then utilized k-means clustering to identify underlying patterns.
As illustrated in Figure~\ref{fig-image-emb}, the clusters can reflect geographic structures: some clusters correspond to spatial proximity and geographically adjacent regions, while others bring together locations that are spatially distant yet share similar attributes. 
For instance, Cluster 1 predominantly contains images from China.
Cluster 2 groups images associated with coastal scenes and Cluster 6 is characterized by mountainous terrain. 
Cluster 4 clusters images depicting culturally and historically significant architecture. 
Incorporating concept-aware mechanism shapes image embeddings toward capturing both spatial contiguity and geographic similarity.

\subsection{Ablation Studies}

\subsubsection{Effectness of Concept Set}
To assess the effectiveness of our proposed Geography-Driven Concept Set, we conduct an ablation study comparing it against a general concept set constructed using the SpLiCE method~\cite{bhalla2024interpreting} on IM2GPS3k dataset. SpLiCE leverages the semantic structure of CLIP’s latent space to decompose representations into sparse, human-interpretable concepts, offering a task-agnostic approach to enhance interpretability without requiring additional training. 

As presented in Table~\ref{tab:ablation_concept}, the results demonstrate the superiority of the Geography-Driven Concept Set across all spatial granularities. Specifically, our tailored concept set achieves accuracy improvements of 1.32\% at 1km levels, compared to the SpLiCE-generated general concept set. And our method achieve improvement in all level ranges,  highlighting the ability of our concept set to capture semantically rich geographic knowledge critical for mid-scale localization tasks. The consistent performance improvements underscore the importance of incorporating geography-specific concepts, which enhance the model’s understanding of world knowledge and its semantic interpretability.

\begin{table}[ht]
\centering
\caption{Comparison of geo-localization accuracy on IM2GPS3k across spatial granularities for models using the Geography-Driven Concept Set versus the SpLiCE-Generated Concept Set.}
\label{tab:ablation_concept}
\small
\setlength{\tabcolsep}{3pt}   
\begin{tabular}{@{}lccccc@{}}
\toprule
\textbf{Concept Set} & \textbf{Street} & \textbf{City} & \textbf{Region} & \textbf{Country}  & \textbf{Continent} \\
 & \textit{(1 km)} & \textit{(25 km)} & \textit{(200 km)} & \textit{(750 km)} & \textit{(2500 km)} \\
\midrule
SpLiCE- & 11.88 & 32.92 & 33.16 & 66.73 & 80.41 \\
Generated & & & & & \\
Geography- & \textbf{13.2} & \textbf{34.0} & \textbf{49.8} & \textbf{68.2} & \textbf{83.5} \\
Driven & & & & & \\
\bottomrule
\end{tabular}
\end{table}

\subsubsection{Effectness of Concept-Aware Alignment Module}
To evaluate the effectiveness of our proposed Concept-Aware Alignment Module, we conducted an ablation study comparing the geo-localization performance of our model with and without this module. The experiments were performed on IM2GPS3k dataset.

As presented in Table~\ref{tab:ablation_module}, the results demonstrate the contribution of the Concept-Aware Alignment Module to geo-localization performance, which improves geo-localization accuracy, achieving 13.2\% and 83.5\% at 1km and 2500km levels, respectively. This improvement aligns with our goal of addressing the loss of world knowledge, as the module enables the model to leverage semantically grounded geographic concepts, thereby improving both accuracy and interpretability. Notably, the significant performance boost at the 1km (22.22\%) and 25km (9.32\%) levels suggests that the module excels in capturing fine-grained semantic cues essential for precise localization. 

\begin{table}[ht]
\centering
\caption{Geo-localization accuracy on IM2GPS3k with and without the Concept-Aware Alignment Module across different spatial granularities.}
\label{tab:ablation_module}
\small
\setlength{\tabcolsep}{1.5pt}   
\begin{tabular}{@{}lccccc@{}}
\toprule
\textbf{Method} & \textbf{Street} & \textbf{City} & \textbf{Region} & \textbf{Country}  & \textbf{Continent} \\
 & \textit{(1 km)} & \textit{(25 km)} & \textit{(200 km)} & \textit{(750 km)} & \textit{(2500 km)} \\
\midrule
W/o  Concept-Aware & 10.8 & 31.1 & 48.7 & 67.6 & 83.2 \\
Alignment Module & & & & & \\
W/ Concept-Aware & \textbf{13.2} & \textbf{34.0} & \textbf{49.8} & \textbf{68.2} & \textbf{83.5} \\
 Alignment Module & & & & & \\
\bottomrule
\end{tabular}
\end{table}

\section{Discussion}

\begin{figure}[t]
\centering
\includegraphics[width=0.9\columnwidth]{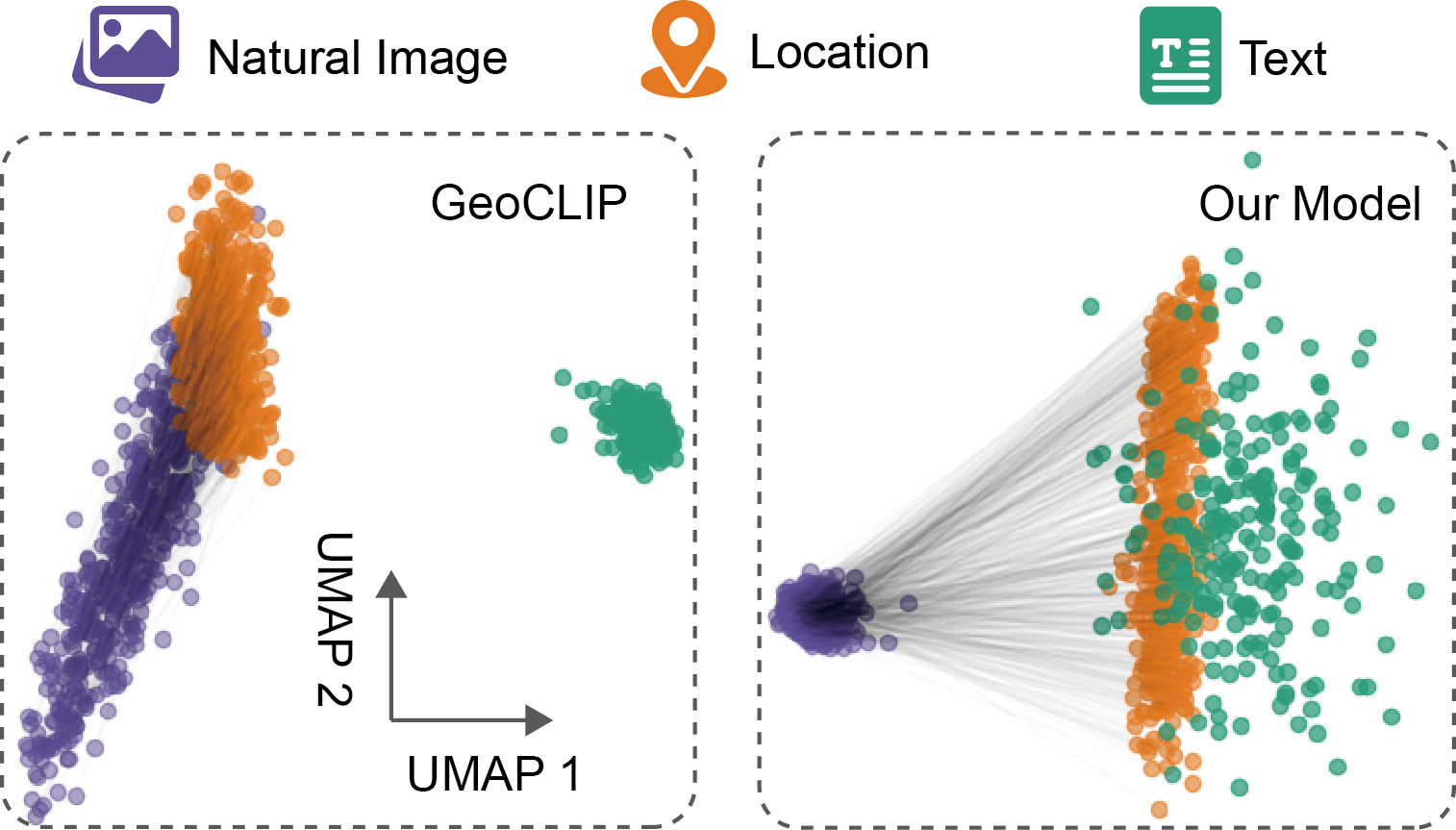} 
\caption{
\textbf{UMAP visualization of embeddings from images, locations, and concepts.}
We project image–location embedding pairs and concept text embeddings (from CLIP) into 2D using UMAP, with lines connecting each image to its corresponding location. Compared to GeoCLIP, our model exhibits a more structured and semantically coherent embedding space.
}
\label{modality-gap}
\end{figure}

To investigate the impact of incorporating concepts into the image-GPS contrastive representation learning framework from a multimodal perspective, we visualize UMAP projections of embeddings from GeoCLIP and our model, derived from image, location, and text modalities. Text embeddings use CLIP-generated representations of geographic concepts employed in training. Figure~\ref{modality-gap} reveals a significant “modality gap”~\cite{liang2022mind} in GeoCLIP, where image and location embeddings are distant from text embeddings, likely due to its image-GPS alignment suppressing semantic alignment. Conversely, our model’s embeddings show cohesive alignment across modalities, with geographic concepts centrally aligned, enhancing interpretability and geo-localization performance.

\section{Conclusion}
In this paper, we presents an interpretable geo-localization framework that integrates geographic concepts into contrastive learning for image-GPS alignment. 
We propose a Geography-Driven Concept Set tailored for human-understandable semantic concepts, and a Concept-Aware Alignment Module that enhances image and location embeddings with a learnable concept subspace, achieving superior alignment and robust world knowledge integration.
Experiments on Im2GPS3k and downstream geospatial tasks show improved accuracy (e.g., +2.4\% at 1 km over GeoCLIP) and interpretability, with learned explanations aligning closely with real-world geographic patterns and human cognition.


\section{Acknowledgments}
Furong Jia would like to thank Gezhi Xiu for his insightful discussions and valuable guidance.

\bibliography{aaai2026}

\end{document}